\documentclass[%
 reprint,
superscriptaddress,
nofootinbib,
 amsmath,amssymb,
 aps,
floatfix,
]{revtex4-1}

\usepackage{silence}% silencing certain warnings where needed
\usepackage{graphicx}% Include figure files
\usepackage{dcolumn}% Align table columns on decimal point
\usepackage{bm}% bold math
\usepackage{hyperref}% add hypertext capabilities
\usepackage[normalem]{ulem}
\usepackage{enumerate}
\usepackage{xcolor}
\usepackage{forest}
\usepackage{tabularx}
\hypersetup{%
    colorlinks=false,% hyperlinks will be black
    linkbordercolor=red,% hyperlink borders will be red
    pdfborderstyle={/S/U/W 1}% border style will be underline of width 1pt
}

\usepackage{todonotes}
\usepackage{lineno}

\usepackage{ifpdf}
\usepackage{amsmath,amssymb}
\usepackage{latexsym}
\usepackage{subfig}
\usepackage[font=footnotesize]{caption}
\usepackage{tabularx}
\usepackage{amsmath}
\usepackage{multirow}
\usepackage{booktabs}
\usepackage{algorithm}
\usepackage{algorithmic}
\usepackage{rotating}
\usepackage{setspace}
\usepackage{amsthm}
\usepackage{amsfonts}
\usepackage{threeparttable}

\usepackage{color}
\begin{document}

\title{Iterative Genetic Improvement: Scaling Stochastic Program Synthesis}
%\thanks{To be published as TBD}%

\author{Yuan Yuan}
%\thanks{Now at: School of Computer Science and Engineering, Beihang University, Beijing, 100191, China}
\affiliation{Department of Computer Science and Engineering, Michigan State University, East Lansing, MI 48864, USA}
%\affiliation{BEACON Center for the Study of Evolution, Michigan State University, East Lansing, MI, USA}%
\affiliation{School of Computer Science and Engineering, \\ Beihang University, Beijing, 100191, China }

\author{Wolfgang Banzhaf}
\affiliation{Department of Computer Science and Engineering, Michigan State University, East Lansing, MI 48864, USA}
\affiliation{BEACON Center for the Study of Evolution, Michigan State University, East Lansing, MI, USA}%

%\collaboration{CLEO Collaboration}%\noaffiliation
\vspace{1cm}

\date{\today}% It is always \today, today,
             %  but any date may be explicitly specified

\begin{abstract}
Program synthesis aims to {\it automatically} find programs from an underlying programming language 
that satisfy a given specification. %, which has the potential to revolutionize computing. 
While this has the potential to revolutionize computing, 
how to search over the vast space of programs efficiently is an unsolved challenge in 
program synthesis. In cases where large programs are required for a solution, it is generally believed that {\it stochastic} search has advantages over other classes of search techniques. Unfortunately, existing stochastic program synthesizers do not meet this expectation very well,
suffering from the scalability issue. Here we propose a new framework for 
stochastic program synthesis, called iterative genetic improvement to overcome this problem, a technique inspired by the practice of the software development process. 
The key idea of iterative genetic improvement is to apply genetic improvement  to improve a current reference program, and then iteratively replace the reference program by the best program found. 
Compared to traditional stochastic synthesis approaches, 
iterative genetic improvement can build up the complexity of programs incrementally in a more robust way. 
We evaluate the approach on two program synthesis domains: list manipulation and string transformation. 
Our empirical results indicate that this method has considerable advantages over 
several representative stochastic program synthesizer techniques, both in terms of scalability and of solution quality. 
\end{abstract}

\maketitle

\setcounter{figure}{0}

\section{INTRODUCTION}
\label{sec:Introduction}

Program synthesis is a longstanding challenge in artificial intelligence (AI) and 
has even been considered by some as the ''holy grail of computer science'' \cite{gulwani2017program}. 
The goal of program synthesis is to {\it automatically} write a program that has a behavior consistent with a specification. 
The specification itself can be expressed in various forms such as logical specification \cite{manna1980deductive, srivastava2010program}, 
examples \cite{bauer1979programming, gulwani2011automating, gulwani2016programming, balog2017deepcoder} or in natural language descriptions \cite{desai2016program,yaghmazadeh2017sqlizer,shin2019program}. 
Program synthesis techniques have been used successfully in many real-world application domains including data wrangling \cite{gulwani2011automating,devlin2017robustfill, chen2021spreadsheetcoder}, 
program repair \cite{le2012systematic, yuan2020arja}, computer graphics \cite{gulwani2011synthesizing,ellis2018learning} and others. 
Furthermore, viewing machine learning tasks as program synthesis \cite{ellis2015unsupervised, trivedi2021learning} can potentially address some of the difficulties of modern deep learning approaches, e.g., data hunger or poor interpretability, 
leading to more reliable and interpretable AI. 

Recently, program synthesis has seen a renaissance in several different research communities, particularly in the programming language and 
the machine learning community. 
Most research efforts focus on developing more effective search techniques since program synthesis
is a notoriously difficult combinatorial search problem. 
Enumerative search-based synthesis \cite{alur2018search} enumerates programs in the search space according to a specific order, 
a topic well studied in the literature. But this class of techniques is usually inefficient 
and needs to be augmented with strategies (i) to prune the search space \cite{gulwani2011automating, alur2017scaling}, (ii)
to bias the search using probabilistic models \cite{balog2017deepcoder, lee2018accelerating,odena2021bustle}, or (iii) to split a large problem via divide-and-conquer strategies \cite{alur2017scaling,huang2020reconciling}. 
Despite such augmentations, enumerative search still struggles to scale to large program sizes, 
as the search space grows exponentially with program size. 
Another popular class of search techniques is to reduce the program synthesis problem by constraint solving,
and leverage off-the-shelf SAT/SMT solvers to efficiently explore the search space \cite{solar2008program,jha2010oracle,feng2018program}. 
Although this class of techniques has achieved impressive results \cite{alur2019syguscomp}, it also has difficulty in
synthesizing large programs due to the limited power of the underlying SAT/SMT solvers. 

Stochastic program synthesis (SPS) employs \emph{stochastic search methods} such as the Metropolis-Hastings (MH) algorithm \cite{chib1995understanding} or genetic programming (GP) \cite{koza1992genetic, banzhaf1998genetic} in order to explore the 
space of programs. 
Compared to the above mentioned search techniques, SPS
is a very promising technique for addressing harder synthesis problems that require larger programs \cite{gulwani2016technical}. 
However, this potential of SPS is currently far from being fully exploited.  
According to the experiments done in \cite{alur2013syntax}, a typical MH approach for program synthesis has been shown to be
not very effective compared to other approaches. 
One possible drawback of the MH approach is that it cannot make the corresponding local changes 
when a program is close to being correct, because the proposal distribution used can only lead to big changes in a program \cite{sps2018}. 

The Genetic Programming (GP) method was always intended for automatic programming, but has suffered from scalability issues for quite a long time \cite{o2019automatic}. 
For hard problems, GP can be very slow and the well-known continued growth of programs without fitness improvement ('bloat') greatly limits its applicability \cite{gustafson2004problem}. 
Although there have been some methods to handle code growth \cite{luke2006comparison}, 
they usually result in unsatisfactory
or even worse performance \cite{gustafson2004problem, luke2006comparison}. 
Besides MH and GP, stochastic local search (SLS) \cite{hoos2004stochastic}, such as simulated annealing, 
would be another alternative, but it has received little attention \cite{o1995analysis} in program synthesis. 
One serious limitation of SLS is that it can easily get trapped in local minima, given that
the search space of programs is often highly rugged and contains many plateaus \cite{schulte2014software,langdon2017software}. 

In this paper, we propose an \emph{iterative genetic improvement} (IGI) method to make SPS more scalable for finding large programs as solutions. 
Our approach is inspired by a practical software development technique called iterative enhancement \cite{basil1975iterative}, 
where human programmers write a simple initial implementation and then enhance 
it iteratively until a final implementation is achieved. 
Based on this paradigm, our basic idea is to consider a sequence of program improvements just like the evolution from a primitive 
cell to a sophisticated and specialized cell \cite{banzhaf2018some}. 
%To be more specific, 
Our proposed IGI starts with a random program, then the 
current program is improved iteratively by applying genetic improvement (GI) \cite{petke2018genetic}
which evolves modifications to the current program. When no improvements can be made any more by GI, 
a perturbation operator is applied that will allow to continue the iterative improvement process.
Because IGI carefully rewrites small parts of a current program via GI in each iteration, 
it can largely avoid unnecessarily big code changes like those of the MH approach \cite{alur2013syntax}, or the 
code growth problem GP faces, leading to faster search. Moreover, IGI %essentially 
considers a neighborhood of the current program that is much larger than SLS based techniques such as SA \cite{o1995analysis}, so it is more able to avoid or escape local minima and plateaus. 
We demonstrate the superiority of IGI on two different program synthesis domains and compare with 
several representative SPS techniques.  
Our experimental results show that IGI can outperform all of the compared techniques by 
a large margin in terms of scalability. 
In addition, our results also indicate that IGI appears to be less prone to overfitting.

The rest of this paper is organized as follows. 
In Section \ref{sec-Preliminaries and Background} we discuss some background material for our study. 
Section \ref{sec-Iterative Genetic Improvement} describes the proposed IGI method in detail. Section \ref{sec-Experiments} presents empirical results while Section \ref{sec-Related Work} briefly introduces other work related to this study. 
Section \ref{sec-Conclusion} summarizes and concludes.

\section{Preliminaries and Background}
\label{sec-Preliminaries and Background}

\subsection{Domain-Specific Language}
\label{sec-Domain-Specific Language}

A domain-specific language (DSL) is a computer language that is specifically created for a particular domain. 
Program synthesis is usually based on a given DSL in order to take a first cut at the space of possible programs. 

A DSL can be expressed as a \emph{primitive set} which includes \emph{functions} and \emph{terminals} assumed to be useful
for solving problems in a specific domain. Consider a toy DSL with the primitive set
$\{ \texttt{ADD}, \texttt{SUB}, \texttt{EQ}, \texttt{ITE}, 0, 1, \text{IN0}, \text{IN1}\}$. 
In this primitive set, \texttt{ADD}, \texttt{SUB}, \texttt{EQ} and \texttt{ITE} are functions explained in Table \ref{tab-toydsl}, 
and the remaining four entities are terminals. Among the four terminals, integers 0 and 1 are constants, and integers IN0 and IN1 are external inputs to the program.

\begin{table}[htbp]
\renewcommand{\arraystretch}{1.0}
\caption{Description of functions in the DSL.}
\label{tab-toydsl} 
\scriptsize \tabcolsep = 4.0pt
\begin{center}
\begin{ruledtabular}
\begin{tabular}{clcl}
    Symbol &   Arguments  & Return Type & Description \\
\colrule
   \texttt{ADD}   & $x, y$: Integer &   Integer     & Return $x+y$.  \\
  \cmidrule{1-4}
   \texttt{SUB}  &  $x, y$: Integer  & Integer     & Return $x-y$.\\
    \cmidrule{1-4}
   \texttt{EQ}    &  $x, y$: Integer  & Boolean                   &   If $x$ is equal to $y$, return true,\\
   		       &&									     &  otherwise return false.  \\
 
     \cmidrule{1-4}
   \texttt{ITE}  & $c$: Boolean;  & Integer &  If $c$ is true, return $x$, \\
   	             & $x, y$: Integer &									     & otherwise return $y$. \\
\end{tabular}
\end{ruledtabular}
\end{center}
\end{table}

\begin{figure}[htbp]
\centering
\includegraphics[scale=0.5]{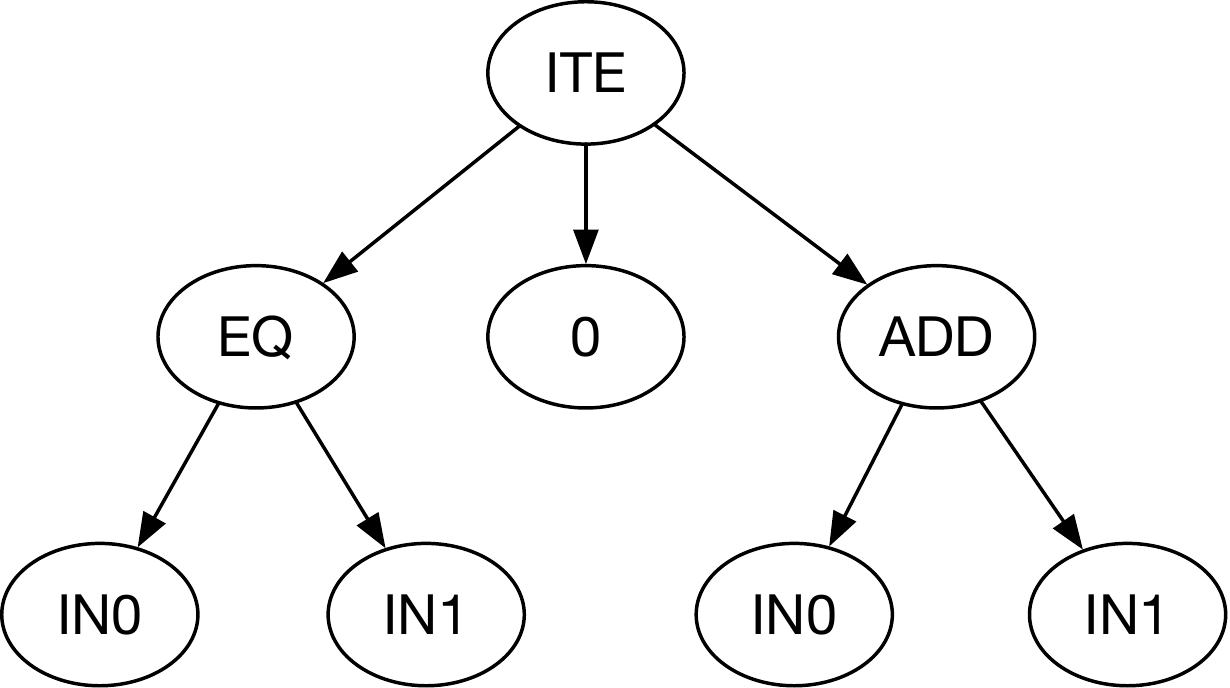}
\caption{The tree of the valid program \texttt{ITE(EQ(IN0, IN1), 0, ADD(IN0, IN1))} in the above toy DSL. } 
\label{fig-tree} 
\end{figure}

The primitive set of a DSL defines the complete hypothesis space of possible programs. Each program can be represented as 
an \emph{expression tree}. Figure \ref{fig-tree} shows such a tree for a valid program in the above toy DSL.

In our experiments, we will consider two DSLs. One is useful for list manipulation (referred to as DSL-LM),
and the other is useful for string transformation (referred to as DSL-ST). They are described in detail in Appendix \ref{sec-DSL for List Manipulation} and Appendix \ref{sec-DSL for String Transformation},
respectively.

\subsection{Programming by Example}
\label{sec-Programming by Example}

Programming by example (PBE) is a  subfield of program synthesis, where
the specification is given in the form of input-output (IO) examples. 

Assume that we have a DSL that defines the space of 
all possible programs denoted by $\mathcal{P}$. In PBE, a task is described by
a set of IO examples $\mathcal{X}=\{(I_{1}, O_{1}), \ldots, (I_{n}, O_{n})\}$. 
We can then say we have solved this task if we find a program $\mathbf{p} \in \mathcal{P}$
that can map correctly every input in $\mathcal{X}$ to the corresponding output, i.e., 
$\mathbf{p}(I_{i})=O_{i}, \forall i = 1, 2, \ldots, n$. Table \ref{tab-io} shows a PBE task with four IO examples and a potential correct program from the toy DSL
described in Section \ref{sec-Domain-Specific Language}.

\begin{table}[htbp]
\renewcommand{\arraystretch}{1.0}
\caption{A PBE task with four IO examples (each input contains two integers) and a possible program in the toy DSL that satisfies all of the four examples.}
\label{tab-io} 
\scriptsize 
\tabcolsep = 5.0pt
\begin{center}
\begin{ruledtabular}
\begin{tabular}{ccl}
%\toprule
 Input ($I_{i}$)&   Output ($O_{i}$)  & Program \\
\colrule
  3, 3   &   0  &   \texttt{ITE(EQ(IN0, IN1), 0, ADD(IN0, IN1))} \\
  2, 5  &     7 \\
  8, 1  &  9   \\
  6, 6   & 0 \\

%\bottomrule
\end{tabular}
\end{ruledtabular}
\end{center}
\end{table}

Our goal is to build a program synthesizer which can find such a program $\mathbf{p}$ in $\mathcal{P}$
according to the IO specification $\mathcal{X}$ within a certain time limit.

\subsection{Genetic Improvement}
\label{sec-Genetic Improvement}

Genetic improvement (GI) \cite{petke2018genetic} is the use of an automated search algorithm, genetic programming, to improve an existing program. 
GI typically conducts the search over the space of \emph{patches}, where a patch constitutes 
a sequence of edits that are applied to the original program and corresponds to a modified program. 
Compared to the original program, an improved version needs to have better 
functional properties (e.g., by eliminating buggy behavior) \cite{le2012systematic, yuan2020arja, yuan2020toward}
or better non-functional properties (e.g., shorter execution time or smaller energy consumption) \cite{langdon2015optimizing, bruce2015reducing}, 
depending on the application scenario or an user's goal. 
GI has achieved notable success in software repair and optimization. 
For example, a GI approach \cite{langdon2015optimizing} can find code that is 70 times faster (on average), when 
applied to a widely-used DNA sequencing system called Bowtie2. 

In this paper, GI improves a program with respect to a fitness function 
measuring how well the program performs over a given set of input-output examples.

%\subsection{Motivating Example}
%\label{sec-Motivating Example}

\section{Iterative Genetic Improvement}
\label{sec-Iterative Genetic Improvement}

\subsection{Overview}
\label{sec-Overview}

The IGI framework proposed here is described in Algorithm \ref{alg-framework}.
First, in Step 1, we create $K$ program trees randomly using ramped half-and-half initialization 
\cite{koza1992genetic} with a depth range of $[d_{\min}, d_{\max}]$. The best among these $K$
random programs is set to become the initial program $\mathbf{p}_{0}$.
This step is intended to find a good starting point for the search by sampling from a (substantial) number of 
programs instead of a single one. 

\begin{algorithm}[H]
\caption{Framework of the Proposed IGI} \label{alg-framework}
  \algsetup{ linenosize=\small, linenodelimiter=.  }
\begin{algorithmic}[1]
\STATE  $\mathbf{p}_{0} \leftarrow \texttt{InitProg}(d_{\min}, d_{\max}, K)$
\STATE $\mathbf{p} \leftarrow \texttt{IterGenImprov}(\mathbf{p}_{0})$
\WHILE {termination criterion is not satisfied}
\STATE $\mathbf{p}'  \leftarrow  \texttt{Perturbation}(\mathbf{p}$)
\STATE $\mathbf{p}'' \leftarrow \texttt{IterGenImprov}(\mathbf{p}')$
\STATE $ \mathbf{p} \leftarrow \texttt{AcceptanceCriterion}(\mathbf{p}, \mathbf{p}'')$
\ENDWHILE
\RETURN the best program found
\end{algorithmic}
\end{algorithm}

\begin{figure}[htbp]
\centering
        \includegraphics[scale=0.22]{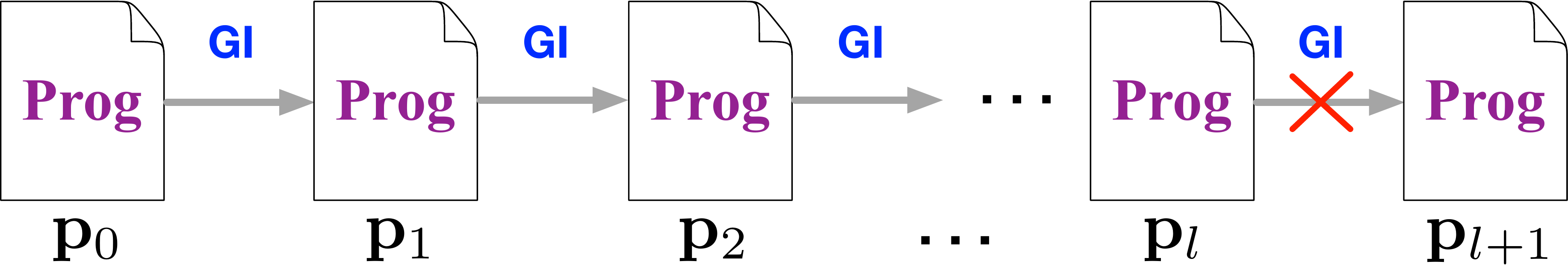}
\caption{
Illustration of Step 2 in Algorithm \ref{alg-framework}. $\mathbf{p}_{i+1}$
is an improved version of $\mathbf{p}_{i}$ for $i=0,1,\ldots, l-1$, that is obtained by applying GI to $\mathbf{p}_{i}$.
Since $\mathbf{p}_{l+1}$ is not better than $\mathbf{p}_{l}$ (i.e., GI fails to improve $\mathbf{p}_{l}$), 
$\mathbf{p}_{l}$ is finally returned by \texttt{IterGenImprov}. 
 } \label{fig-IGI} \end{figure}

In Step 2, we make incremental improvements from 
$\mathbf{p}_{0}$ by applying the GI procedure \emph{iteratively} until 
a program $\mathbf{p}$ is reached that cannot be further improved by the adopted GI. 
In each iteration GI tries to produce an improved version $p_{i+1}$ by searching for 
modifications to the current program $p_{i}$.
%\textcolor{red}{WB: I think this needs to be explained in more detail! How many iterations is GI applied, etc.}
Figure \ref{fig-IGI} illustrates this process where $\mathbf{p}_{l}=\mathbf{p}$, and we call 
the process from $\mathbf{p}_{i}$ to $\mathbf{p}_{i+1}$ an \emph{epoch}, 
where $\mathbf{p}_{i+1}$
is an improved version of $\mathbf{p}_{i}$ obtained through GI. 
 
After Step 2, to continue the search, we need to generate a new starting program for \texttt{IterGenImprov}. 
To do this, a naive strategy is to randomly produce a new program as the 
starting program. However this strategy is obviously not efficient because 
the search history is completely discarded. Here we follow the basic idea
of iterated local search \cite{lourencco2019iterated}. That is, we apply some perturbation
operator to $\mathbf{p}$ that leads to an intermediate program $\mathbf{p}'$ (Step 4 in Algorithm \ref{alg-framework}). 
Then \texttt{IterGenImprov} restarts the search from
$\mathbf{p}'$ and returns $\mathbf{p}''$ (Step 5 in Algorithm \ref{alg-framework}). 
In Step 6, \texttt{AcceptanceCriterion} will decide to which program the next time 
\texttt{Perturbation} is applied. 
In this paper, this criterion just simply returns the better of 
$\mathbf{p}$ and $\mathbf{p}''$.
Steps 4--6 are iterated until some termination criterion is met. 

As can be seen in IGI, we need to compare the quality of two
programs frequently. This is aided by a predefined \emph{fitness function} which can measure 
how well a program satisfies the given specification. Suppose we are given a set of $n$
input-output examples $\{(I_{1}, O_{1}), \ldots, (I_{n}, O_{n})\}$, the fitness function of 
a program $\mathbf{p}$ can be defined as 
$fitness(\mathbf{p}) = \frac{1}{n}\sum_{i=1}^{n} sim(O_{i}, \mathbf{p}(I_{i}))$, 
where $sim$ function indicates the similarity between the expected output
$O_{i}$ and the actual output $\mathbf{p}(I_{i})$. 
In our study, 
%for the list manipulation domain, 
for DSL-LM,
$sim$ returns 1 if $O_{i}=\mathbf{p}(I_{i})$ otherwise returns 0. 
%As for the string transformation domain, 
As for DSL-ST, 
we use a finer-grained $sim$ function 
which returns the normalized Levenshtein similarity\footnote{We use the \texttt{levenshtein.normalized\_similarity} function from  https://github.com/life4/textdistance, calculating a value between $[0, 1]$, with 1 returned if the two strings are the same.}
between two strings. In our approach, a program with larger fitness is better, while in the case of two programs having
the same fitness, the program with smaller size is deemed better according to Occam's razor.
A program $\mathbf{p}$ will satisfy the given specification iff it achieves maximum fitness (i.e., $fitness(\mathbf{p})=1$). 

Section \ref{sec-Applying GI} will detail how to apply GI to the current program $\mathbf{p}_{i}$ to get an 
improved version $\mathbf{p}_{i+1}$, which corresponds to one epoch in \texttt{IterGenImprov} (see Figure \ref{fig-IGI}).
Section \ref{sec-Perturbation Operator} will explain how to conduct the perturbation operator, which corresponds to 
Step 4 in Algorithm \ref{alg-framework}.

\subsection{Applying Genetic Improvement}
\label{sec-Applying GI}

\subsubsection{Patch Representation}
\label{sec-Patch Representation}

A program different from the program we want to improve can be coded as the difference between the original and the new program. In computing, this is known as a program patch, represented as a \emph{sequence} of edits to the program's expression tree. 
In this patch representation, we define three kinds of edits: \emph{replacement}, \emph{insertion} and \emph{deletion}. 
Syntactic and type constraints are considered in these edits in order to ensure that the modified tree remains legal.
To randomly generate an edit, we first select a node randomly in the expression tree that is called target node. 
Then we choose one of the three kinds of edits randomly.  

\begin{figure}[htbp]
\centering
        \includegraphics[scale=0.3]{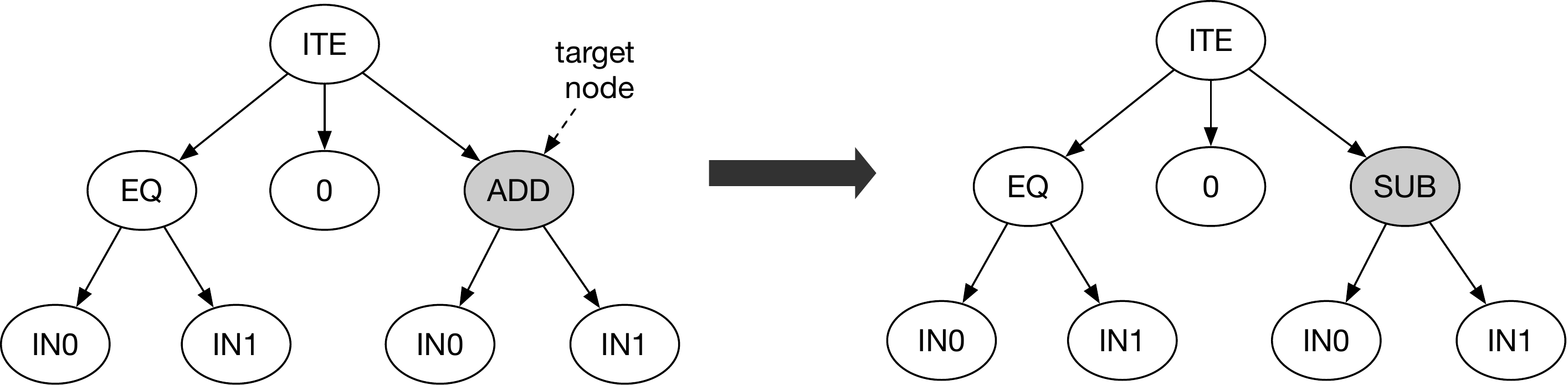}
\caption{Illustration of a replacement edit (shaded nodes are changed). 
 } \label{fig-replace} 
\end{figure}

1) \textbf{Replacement}: If we choose to perform replacement, the target node is replaced by 
another random primitive from the primitive set which has the same number of arguments, the same argument types and the same return type. 
This is illustrated in Figure \ref{fig-replace}. 

\begin{figure}[htbp]
\centering
        \includegraphics[scale=0.29]{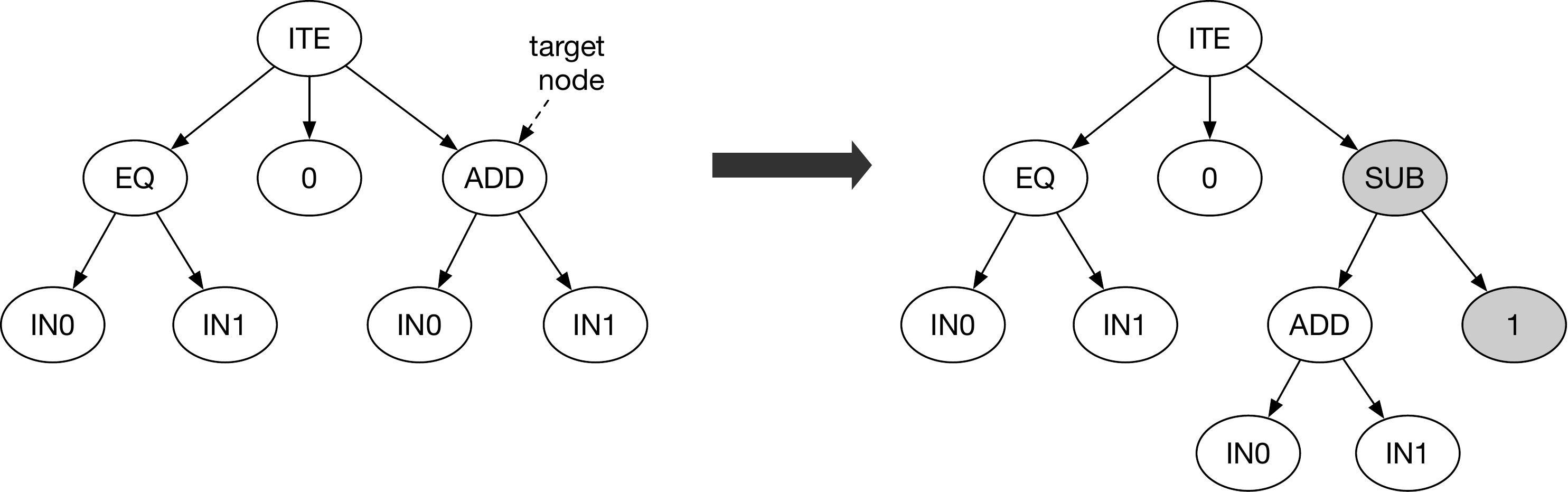}
\caption{Illustration of an insertion edit (shaded nodes are the newly inserted nodes). 
 } \label{fig-insert} 
\end{figure}

2) \textbf{Insertion}: If we choose to perform insertion, for the primitive that can be inserted, 
its return type and at least one of its argument types should be the same as the return type of the target node.
Such a primitive is selected at random from the primitive set. Then this primitive will replace the target node, and 
the subtree rooted at the target node will become one of its child tree that requires the same data type. All the remaining children 
of this primitive will be selected randomly from the set of terminals with the corresponding data types. This is illustrated in Figure \ref{fig-insert}. 

\begin{figure}[htbp]
\centering
        \includegraphics[scale=0.3]{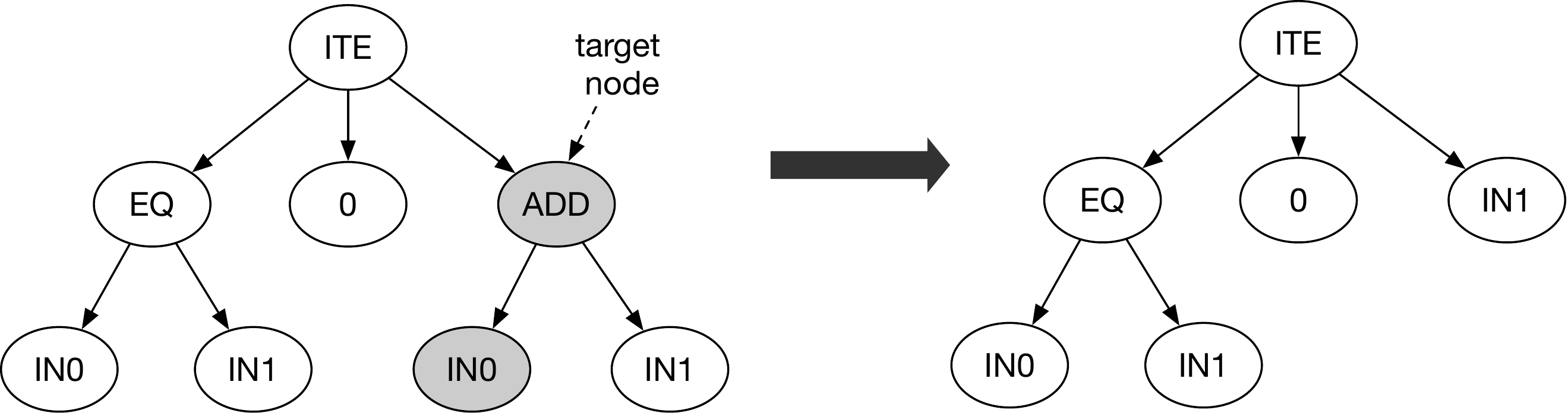}
\caption{Illustration of a deletion edit (shaded nodes are removed from the original tree). 
 } \label{fig-delete} 
\end{figure}

3) \textbf{Deletion}: If we choose to perform deletion, one node is randomly 
chosen from the children of the target node that have the same return type as the target node. 
The subtree rooted at this node will replace the subtree rooted at the target node. 
This is illustrated in Figure \ref{fig-delete}.

\begin{figure}[htbp]
\centering
        \includegraphics[scale=0.29]{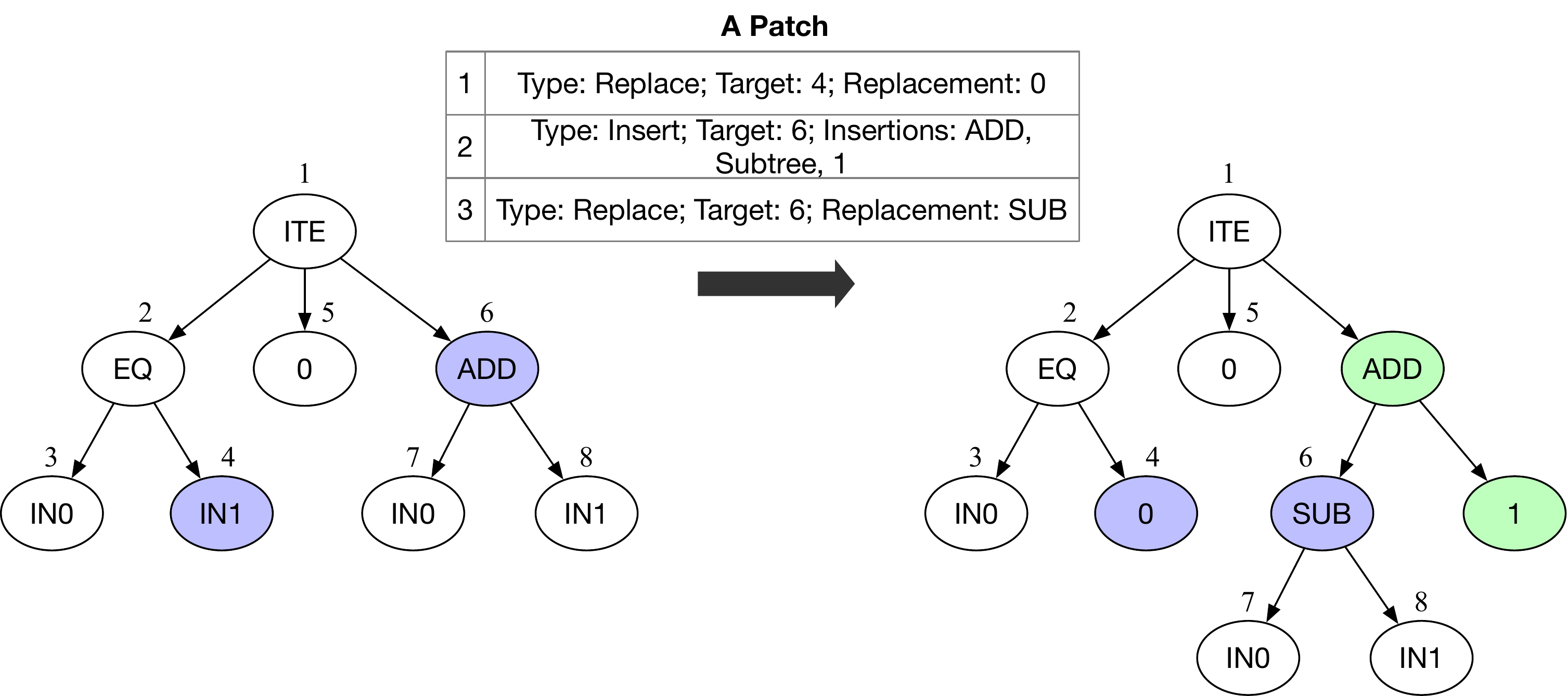}
\caption{Illustration of a patch that contains three edits. The number above the node is the ID of the node. 
The blue shaded nodes are changed by replacement edits and the green shaded nodes are the new nodes added by 
the insertion edit. 
 } \label{fig-patch} 
\end{figure}

A patch contains a list of concrete edits and each edit is performed sequentially
when applying the patch. This is illustrated in Figure \ref{fig-patch}.

%A patch is made up of a list of edits, where each edit is performed sequentially. 
%Figure shows the applying of a patch with three edits to 

\begin{figure}[htbp]
\centering
        \includegraphics[scale=0.25]{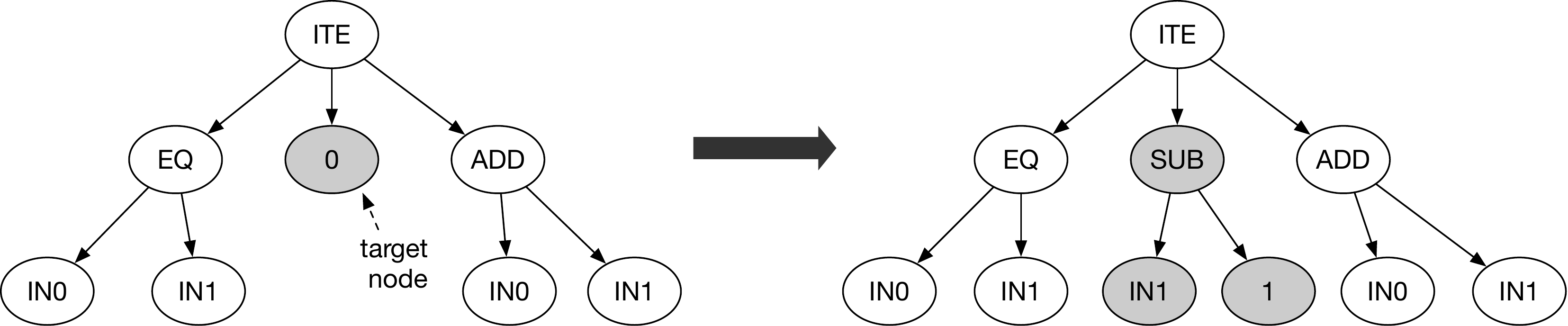}
\caption{Illustration of a replacement edit (shaded node in the original tree is replaced by the shaded nodes in the new tree). 
 } \label{fig-replace2} 
\end{figure}

In our study, we find that the above replacement/insertion edits sometimes
struggle to introduce the following two kinds of primitives into the program. 
First, if the return type of a primitive is different from all its 
argument types (e.g., the primitive \texttt{LEN} in the DSL-ST), we 
know that it is hard or even impossible to
bring the two kinds of primitives using the above insertion edit.
Second, if there are no corresponding terminals for some of its argument types of a primitive
(e.g., the primitive \texttt{Take} in the DSL-LM), the situation is similar.
Also, a replacement edit might have little chance to get introduced into a program, 
if there are few (or even no) primitives with the same return type and argument types in
the set. Should the desired program really require this primitive, the search will become inefficient. 
We address the generation of replacement and insertion edits to introduce the two kinds of primitives 
easily as follows:\\
For the replacement edit, when the target node is a leaf, a function with the same return type is also allowed to replace the 
target node and the arguments of this function will be filled with random terminals having the 
corresponding types. This is demonstrated in Figure \ref{fig-replace2}. 

\begin{figure}[htbp]
\centering
        \includegraphics[scale=0.24]{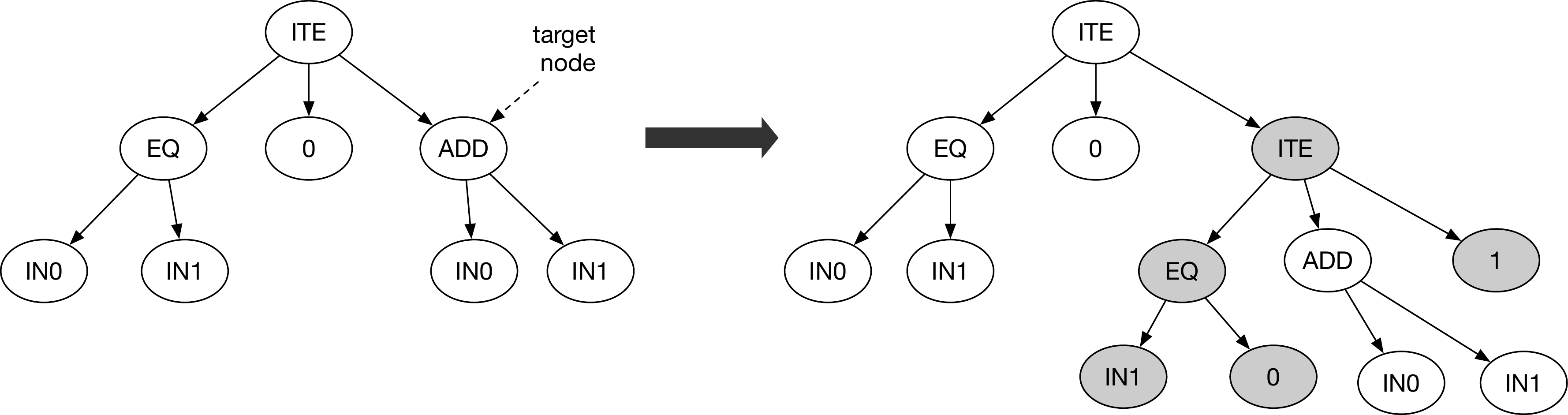}
\caption{Illustration of an insertion edit (shaded node are the newly inserted nodes). 
 } \label{fig-insert2} 
\end{figure}

As for the insertion edit, when the algorithm fails to find a terminal for a child of the inserted node, it instead chooses
a function with the desired return type and fill its arguments with random terminals having the corresponding types. 
This is demonstrated in Figure \ref{fig-insert2}.

Note that there may exist conflicts between two edits. 
For example, a replacement edit as shown in Figure \ref{fig-replace} and 
a deletion edit as shown in Figure \ref{fig-delete} cannot take effect at the same time. 
We resolve such conflicts at the time of generating or applying a patch by disabling 
the latter one if it conflicts with a previous edit.

Using this patch representation, we can search for patches 
to the current program $\mathbf{p}_{i}$ that are able to produce an improved 
version $\mathbf{p}_{i+1}$. 
In our framework, we provide two alternative search 
algorithms: stochastic beam search (SBS) and linear genetic programming (LGP) \cite{brameier2007linear}, 
which are described in the next subsections
 \ref{sec-Stochastic Beam Search} and \ref{sec-Linear Genetic Programming}, respectively.

\subsubsection{Stochastic Beam Search}
\label{sec-Stochastic Beam Search}

In stochastic beam search (SBS), the search for  patches of length $L+1$ is
based on the patches of length $L$ (i.e., there are already $L$ edits in the patch). 
Suppose that we currently have a set of $B$ patches of length $L$.
Then for each of the $B$ patches, we produce $C$ copies
and append a randomly generated edit to each copy where 
the appended edit should not conflict with the previous $L$ edits. 
Having produced a total of $B\times C$ patches of length $L+1$, 
tournament selection is used to select $B$ patches from the total $B\times C$. 
Based on these $B$ patches of length $L+1$, SBS continues to explore further
patches, this time of length $L+2$.

SBS starts the search from $B$ empty patches (i.e., length $L=0$), with a 
limiting parameter $L_{\max}$ used to restrict the maximum length of allowed patches. 
Once SBS finds a program $\mathbf{p}_{i}'$ that is better than the current program $\mathbf{p}_{i}$, 
we need to decide whether to continue SBS or just return $\mathbf{p}_{i}'$ as $\mathbf{p}_{i+1}$
for the next GI epoch. In this study, our strategy is that if $\mathbf{p}_{i}'$ is the 
best program ever found from the beginning of IGI, SBS in this GI epoch is deemed to be fruitful and 
continues searching for better return programs until it reaches $L_{\max}$.
Otherwise $\mathbf{p}_{i}'$ can be simply returned as $\mathbf{p}_{i+1}$. 
Note that SBS may fail to find any improved version of the current program 
$\mathbf{p}_{i}$ after reaching $L_{\max}$, in this case the perturbation operator will be invoked.

\subsubsection{Linear Genetic Programming}
\label{sec-Linear Genetic Programming}

In the linear genetic programming (LGP) approach, 
the search starts with a population of $N$ random patches denoted $\mathcal{P}_{0}$.
To produce patches with diverse lengths,  
each patch length in $\mathcal{P}_{0}$ is drawn from a  $1+Poisson(1)$ distribution, 
where $Poisson(\lambda)$ is a Poisson distribution with parameter $\lambda=1$. 

In the $g$-th generation of LGP, we use tournament selection to select two 
parent patches from $\mathcal{P}_{g}$ and apply \emph{crossover} and \emph{mutation}
to the two patches to generate two offspring. 
By repeating this process $N/2$ times, $N$ offspring patches are generated which 
constitute the next population $\mathcal{P}_{g+1}$. 

One-point crossover between two parent patches is used. Suppose that the length of 
the two parents is $L_{1}$ and $L_{2}$, respectively. We use a cut point in the two parents 
given by $\lfloor\alpha L_{1}\rfloor$ and $\lfloor\alpha L_{2}\rfloor$, where
$\alpha \in (0,1)$ is a random value. The edits after the cut points are swapped 
between the two parent patches, leading to two offspring.  

Mutation is then applied to each of the two offspring patches. 
When applying mutation to a patch, an edit in the patch 
is selected uniformly at random, and with equal probabilities 
removed, replaced with a random edit, or
a random edit is inserted after the selected one.

In LGP, we use the same strategy as in SBS to determine when to terminate the search and return 
the best program LGP has found during the current epoch of GI.

\subsection{Perturbation Operator}
\label{sec-Perturbation Operator}

The goal of the perturbation operator is to provide a new good starting 
program for the next application of \texttt{IterGenImprov} in case we get stuck.
% \textcolor{red}{WB: Do we only apply perturbation in case we get stuck, or always, even if the previous epoch has led to an improvement?}

When applying the perturbation to $\mathbf{p}$ (Step 4 in Algorithm \ref{alg-framework}), 
some useful components of $\mathbf{p}$ need to be conserved, in order to 
provide a new good starting program for the next application of \texttt{IterGenImprov}.
At the same time the perturbation should not be too small, in order to avoid staying stuck 
in the same local optimum as the previous epoch of \texttt{IterGenImprov}. 

In our framework, the perturbation operator works as follows:
First, we collect a set of nodes denoted by $\mathcal{N}$ from the expression tree of $\mathbf{p}$, where the size of the subtree rooted at each node
is not smaller than $S_{\min}$. Second, we randomly select a node $\nu$ from $\mathcal{N}$. For convenience, the subtree rooted at node $\nu$ 
is denoted by $\mathcal{T}$ and its size is denoted by $S_{\mathcal{T}}$. 
Third, we randomly generate $M$ trees with size in the range $[1, S_{\mathcal{T}}]$ and the same return type as $\mathcal{T}$. 
Fourth, we replace $\mathcal{T}$ in the tree of $\mathbf{p}$ with each of the $M$ trees in turn, and obtain $M$
new programs. Last, the best program among the $M$ programs is returned as the perturbed program $\mathbf{p}'$. 

In addition, we handle two special cases: One is that $\mathcal{N}=\emptyset$ and the 
other is that the selected node $\nu$ is the root node. 
For both of these cases, we invoke the initialization function \texttt{InitProg}
to produce a program that is used as the perturbed program $\mathbf{p}'$. 

As can be seen from above, $S_{\min}$ is the parameter that  restricts the minimum strength of a perturbation. 
Since we know that a random subtree replacement usually leads to a bad 
fitness of the new program in program synthesis, our perturbation operator generates $M$
candidate program variants instead of a single one.  
Moreover, with a relatively low probability, the perturbation operator can ignore much (when $\nu$ is the node very close to root)  or
even all (when $\nu$ is the root node) of the information of $\mathbf{p}$, 
which can help to explore other promising regions of the program space.

%Note that in a program's pare tree, most of nodes
%are located close to the leaves. 

%our perturbation operator has the following features:
%1) $S_{\min}$ can restrict the minimum strength of perturbation;
%2) Generating $M$ program variants instead of a single one can have more 
%chance to 

%1) $S_{\min}$ can control the strength of the perturbation; 
%2) $M$ can control the quality of the perturbed program; 
%3) 

%As can be seen above, $S_{\min}$ can control the strength of the perturbation. 

%Note that set $\mathcal{N}$ may be empty. In this case, 

\section{Experiments}
\label{sec-Experiments}

\subsection{Experimental Setup}
\label{sec-Experimental Setup}

The IGI framework is instantiated here with SBS (see Section \ref{sec-Stochastic Beam Search}) 
and LGP (see Section \ref{sec-Linear Genetic Programming}), resulting in 
two algorithms IGI-SBS and IGI-LGP, respectively. 
The source code is implemented in Python and has been made available at %\textcolor{red}{WB: I didn't find it.} 
GitHub.\footnote{The source code is available at https://github.com/yyxhdy/igi/} 
All the experiments are conducted on the Intel Xeon E5-2680 2.4GHz CPU processor with 20GB memory.

\subsubsection{Benchmarks}
\label{sec-Benchmarks}

We evaluate the algorithms on synthesis tasks from two different application domains:
list manipulation and string transformation. The corresponding DSLs for the two domains 
are described in Appendix \ref{sec-DSL for List Manipulation} and \ref{sec-DSL for String Transformation}. 

For the list manipulation domain, 200 benchmarks\footnote{These benchmarks are available at https://github.com/yyxhdy/igi/dataset} are generated, similar to \cite{balog2017deepcoder}. In particular, we first randomly generate a program with a number 
of tree nodes in the range between 10 and 15. Then we generate random input/output pairs for this program. 
This process is repeated until we either obtain 100 valid input-output examples
or no examples can be found within a fixed iteration limit, in which case we discard it and start over again.
%In the latter case, we randomly generate a fresh new program, and restart this process of generating input-output examples. 
Note that in order to test the scalability of SPS techniques, we consider larger oracle programs than existing studies \cite{balog2017deepcoder,feng2018program,chen2020program}.
For example, oracle programs with at most seven nodes were investigated in \cite{balog2017deepcoder}. 

As for the second domain, string transformations, we use the dataset consisting of all 
SyGuS tasks 
whose outputs are only strings from the PBE-Strings track in 2018 and 2019 \cite{sygus2014}. This results in 185 tasks in total. 
The semantic specifications consist of 2--400 examples.

\subsubsection{Baseline Algorithms}
\label{sec-Baseline Algorithms}

IGI-SBS and IGI-LGP are compared to the following representative SPS techniques: 

\textbf{MH} \cite{alur2013syntax} ---
This approach is an adaption of the 
algorithm proposed in \cite{schkufza2013stochastic}
for program superoptimization. 
It uses the Metropolis-Hastings procedure to sample programs.
Given the current program $\mathbf{p}$, a variant $\mathbf{p}'$ with the same size is obtained by conducting a random subtree replacement. 
The probability of adopting $\mathbf{p}'$ as the new current program is given by the Metropolis-Hastings acceptance
ratio $\alpha(\mathbf{p},\mathbf{p}')=\min\{ 1, \exp(\beta (C(\mathbf{p}) - C(\mathbf{p}')) \}$,
where $\beta$ is a smoothing constant and $C(\mathbf{p})=\sum_{i=1}^{n} \{ 1- sim(O_{i}, \mathbf{p}(I_{i})) \}$
indicates the degree  to which $\mathbf{p}$ violates a set of given input-output examples. 
The approach starts search for programs of size $k=1$, but 
switches at each step with some probability $p_{m}$ to search for programs of size $k+1$ or $k-1$.

\textbf{GP} \cite{poli2008field} --- 
This is a highly optimized genetic programming system, often referred to as TinyGP. It employs a steady state algorithm
where only one individual in the population is replaced each generation. During program evolution the system uses tournament selection to select mating parents, and subtree crossover and point mutation to generate offspring.

\textbf{SIHC} \cite{o1995analysis} --- 
This algorithm uses stochastic iterated hill climbing (SIHC) to discover
programs. It starts with a random program as the current program $\mathbf{p}_{c}$,  
then a mutation operator called HVL-Mutate is applied to the current program to obtain a
variant $\mathbf{p}_{c}'$. If $\mathbf{p}_{c}'$ is better than $\mathbf{p}_{c}$, $\mathbf{p}_{c}'$
will replace $\mathbf{p}_{c}$ as the new current program and the search will move onward from it. Otherwise, another mutation to 
$\mathbf{p}_{c}$ is tried. The maximum number of mutations that can be applied 
to $\mathbf{p}_{c}$ is given by a parameter $T_{\max}$. 
If none of the $T_{\max}$ variants is better than $\mathbf{p}_{c}$, 
$\mathbf{p}_{c}$ is discarded and a new current program is generated at random.

\textbf{SA} \cite{o1995analysis} --- 
This algorithm uses simulated annealing (SA) to discover programs. 
It is somewhat similar to SIHC, but at each step only a single variant $\mathbf{p}_{c}'$ is
generated from $\mathbf{p}_{c}$ using HVL-Mutate, 
and the variant 
$\mathbf{p}_{c}'$ will be accepted as the new current program 
with probability $\min\{1, \exp((fitness(\mathbf{p}')- fitness(\mathbf{p}))/T_{c}    ) \}$, 
where $T_{c}$ is the current temperature which is decreased by an exponential rate.

Note that the original HVL-Mutate does not consider data-type constraints in a DSL, so it is not 
applicable to synthesis tasks where there are multiple data-types. 
In our experiments, we replace the HVL-Mutate in SIHC and SA with the 
replacement/insertion/deletion edit described in Section \ref{sec-Patch Representation}, in order to 
ensure a fair comparison.

\subsubsection{Parameter Settings}
\label{sec-Parameter Settings}

The timeout for each run of the algorithm on a benchmark is set to one hour in these experiments, and in each run
the algorithm will be terminated at once when a solution program is found. 

The key parameters of all algorithms considered are tuned on two additional difficult 
synthesis tasks,\footnote{The definitions of the two tasks are available at https://github.com/yyxhdy/pt} 
one for each domain. Grid search is used to find the best parameter combination 
among 81 combinations for IGI-SBS, 72 combinations for IGI-LGP,  50 combinations for MH, 81 combinations for GP, 40 combinations for SIHC and 50 combinations for SA. 
Details can be seen in Appendix \ref{sec-Parameter Selection}.

\subsection{Results for List Manipulation}
\label{sec-Results for List Manipulation}

We evaluate IGI-SBS and IGI-LGP on 200 benchmarks from the list manipulation domain 
and compare them with MH, GP, SIHC and SA. Each algorithm runs only once on each benchmark. 
Table \ref{tab-lispres} summarizes the comparison results, where we list for each algorithm 
the number of benchmarks solved (``Total solved''), the number of benchmarks solved with the fastest solving time (``Fastest solved''), 
the number of benchmarks solved with the smallest program size (``Smallest solved''), 
the average and median times to find a solution, 
and the average and median sizes of solution programs. Here program size refers to the number of nodes in the program's expression tree.

\begin{table}[htbp]
\caption{Main result (in the list manipulation domain) comparing the performance of all algorithms considered. 
\label{tab-lispres}} \scriptsize %\tabcolsep = 2.0pt
%\begin{center}
\begin{ruledtabular}
\begin{tabular}{lcccccc}

Statistics & IGI-SBS & IGI-LGP & MH & GP & SIHC & SA \\
\colrule
Total solved 	    							& 141 & 136 & 24 & 65 & 80 & 62 \\
Fastest solved           						& 76 & 51 & 1 & 2 & 22 & 4 \\
Smallest solved            						& 74 & 64 & 6 & 14 & 49 & 14 \\
Average time            							& 581.76 & 596.28 & 474.77 & 726.97 & 615.26 & 1135.35 \\
Median time             							& 229.78 & 173.27 & 28.85 & 404.61 & 198.15 & 884.54 \\
Average size            							& 12.09 & 12.31 & 12.75 & 16.32 & 10.41 & 22.5 \\
Median size             							& 12.0 & 12.0 & 8.5 & 12.0 & 10.0 & 15.5 \\

\end{tabular}
\end{ruledtabular}
%\end{center}
\end{table}

Out of 200 benchmarks, IGI-SBS and IGI-LGP can solve 141 and 136 benchmarks respectively. 
Other algorithms perform much worse than IGI-SBS and IGI-LGP in terms of total number of
benchmarks solved. The most competitive one is SIHC, but it can only solve 80 benchmarks which is just about 
57\% of that by IGI-SBS.
GP and SA solve similar number of benchmarks (65 for GP and 62 for SA).
MH can only solve 24 benchmarks and is significantly outperformed by all other algorithms. 

IGI-SBS and IGI-LGP are the fastest solvers in 76 and 51 benchmarks, respectively. 
Next to IGI-SBS and IGI-LGP, SIHC is the fastest only in 22 benchmarks. 
In terms of average and median times, IGI-SBS and IGI-LGP 
show clear advantages over GP and SA, and show similar performance to SIHC. 
Although both, average and median times consumed by MH are smallest, 
MH scales poorly.

The solution quality can be roughly measured by the size of a solution program \cite{lee2021combining}. 
According to average and median sizes, solutions found by IGI-SBS and IGI-LGP have overall better quality than those found by GP and SA. 
Compared to IGI-SBS and IGI-LGP, SIHC can generate solutions with smaller average and median sizes. 
This is reasonable because IGI-SBS and IGI-LGP can solve many benchmarks that require larger solution 
programs whereas SIHC can not. 
Moreover, IGI-SBS and IGI-LGP can provide the smallest solutions 
for 74 and 64 benchmarks respectively, whereas SIHC can only provide the smallest solutions 
for 49 benchmarks.

Interestingly, SA needs much more time to find a solution compared to 
the other algorithms. In addition, the solutions found by SA  
are also much larger than those by the other algorithms. One possible reason is
that the space of programs contains a series of discrete plateaus.
SA always accepts solutions with the same fitness, so it may 
spend much time to traverse these plateaus. 
Moreover, during this process, many subtree structures with no effect on fitness \cite{banzhaf1996effect}
could be added into the code, producing larger and larger programs. 
Our proposed IGI explores a large neighborhood of the current program
using GI techniques, so it can avoid or escape from plateaus more easily.

\begin{figure}[htbp]
\centering
\subfloat[ ]{\includegraphics[width=0.8\columnwidth]{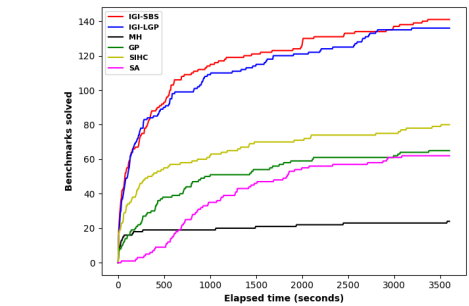}%
\label{fig-lisptime}}
\hspace{20pt}
\subfloat[ ]{\includegraphics[width=0.8\columnwidth]{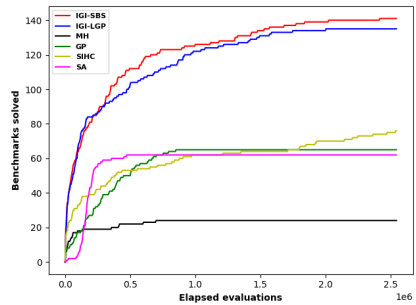}%
\label{fig-lispeval}}
\caption{a) Number of benchmarks solved versus the computation time; (b) Number of benchmarks solved versus the number of evaluations in the list manipulation domain. }
\label{fig-lispte}
\end{figure}

In Figure \ref{fig-lispte}, we plot the number of benchmarks solved versus computation time (Figure \ref{fig-lispte}\subref{fig-lisptime})
and the number of evaluations (Figure \ref{fig-lispte}\subref{fig-lispeval}) for each algorithm. 
Compared to the baselines, the number of benchmarks solved by IGI-SBS or IGI-LGP
increases more steadily with elapsed time/evaluations. 
For MH, almost all the solutions are found during the very early period of 
the search. Considering that MH searches for progressively larger programs, 
this implies that its search mechanism is inadequate for synthesizing 
larger programs. Note that SA gets stuck after a relatively small number of evaluations (i.e., about $5\times10^{5}$), 
but these evaluations cost most of the budget time (i.e., about 3000 seconds). 
This implies that SA tends to examine larger programs whose execution times are longer.

\begin{table}[htbp]
\renewcommand{\arraystretch}{1.0}
\caption{Results for 10 hard benchmarks (in the list manipulation domain) that are randomly chosen. The data indicates the number of successful runs out of 50. 
Best performance is shown in bold. 
}\label{tab-lisprepeat} \scriptsize \tabcolsep = 2.0pt
\begin{center}
\begin{ruledtabular}
\begin{tabular}{cccccccc}
%\toprule
Benchmarks   &  IGI-SBS & IGI-LGP & MH & GP & SIHC & SA \\
   \colrule
   
L15  & 20 & \textbf{24} & 0 & 1 & 0 & 0 \\
L53  & \textbf{20} & 14 & 0 & 0 & 0 & 0 \\
L82  & 4 & \textbf{8} & 0 & 1 & 0 & 7 \\
L129  & \textbf{37} & 36 & 0 & 1 & 0 & 0 \\
L172  & \textbf{15} & 12 & 0 & 0 & 0 & 0 \\
L179  & 29 & \textbf{35} & 0 & 12 & 0 & 10 \\
L180  & 26 & \textbf{32} & 0 & 0 & 4 & 4 \\
L183  & 30 & \textbf{39} & 4 & 13 & 1 & 0 \\
L188  & \textbf{32} & 25 & 0 & 0 & 0 & 0 \\
L197  & \textbf{32} & 28 & 0 & 0 & 0 & 0 \\

%\bottomrule
\end{tabular}
\end{ruledtabular}
\end{center}
\end{table}

Due to the stochastic nature of SPS techniques we measure the performance 
of each technique as the number of runs out of 50 that solve the
benchmark, called success rate. Table \ref{tab-lisprepeat} lists the success rates of each 
algorithm on 10 hard benchmarks. These benchmarks are randomly chosen out of the 200 benchmarks 
and can be solved by no more than three techniques,
according to the results in Table \ref{tab-lispres}.
It can be seen from Table \ref{tab-lisprepeat} that 
IGI-SBS and IGI-LGP perform similarly well and obtain the best success rates 
on some of the 10 benchmarks. IGI-SBS and IGI-LGP 
perform much better than all baselines except on L82 where SA is very competitive. 
On L53, L172, L188 and L197, IGI-SBS and IGI-LGP can achieve decent
success rates, whereas all the baselines never succeed. 

In summary, the IGI algorithms clearly outperform all the baselines in
terms of scalability in the list manipulation domain with overall better solution quality. 
Moreover, they also have an overwhelming advantage in solving hard synthesis tasks.

\subsection{Results for String Transformation}
\label{sec-Results for String Transformation}

Table \ref{tab-sliares} shows the main result in the string transformation domain. 
IGI-SBS and IGI-LGP solve the most number of benchmarks (i.e., 172 for both of them), followed by
GP that solves 159. So IGI algorithms again outperform all the 
other algorithms in terms of scalability. Moreover, they are also superior to others in terms of synthesis time.  
In particular, IGI-SBS finds solutions in an average time of 73.39 seconds, whereas SIHC, the fastest baseline,
consumes 128.16 seconds on average.

\begin{table}[htbp]
\renewcommand{\arraystretch}{1.0}
\caption{Main result (in the string transformation domain) comparing the performance of all algorithms considered. 
}\label{tab-sliares} \scriptsize \tabcolsep = 3.0pt
\begin{center}
\begin{ruledtabular}
\begin{tabular}{lcccccc}
%\toprule

Statistics & IGI-SBS & IGI-LGP & MH & GP & SIHC & SA \\
   \colrule
   
 Total solved 					& 172 & 172 & 108 & 159 & 156 & 158 \\
 Fastest solved					& 63 & 50 & 2 & 5 & 51 & 3 \\
 Smallest solved 				& 56 & 38 & 38 & 14 & 88 & 6 \\
 Average time  					& 73.39 & 109.45 & 360.04 & 135.3 & 128.16 & 325.29 \\
 Median time  					& 4.02 & 5.25 & 29.56 & 39.39 & 6.94 & 91.79 \\
 Average size					& 19.1 & 20.78 & 24.82 & 42.37 & 14.28 & 277.43 \\
Median size 				        & 17.0 & 19.0 & 15.0 & 27.0 & 12.5 & 120.0 \\

%\bottomrule
\end{tabular}
\end{ruledtabular}
\end{center}
\end{table}

In terms of solution size, SA severely suffers from the code growth problem and  average 
solution size reaches over 277, possibly due to the same reason as in Section \ref{sec-Results for List Manipulation}.
GP suffers from a similar problem with average and median sizes of 42.37 and 27. 
IGI-SBS and IGI-LGP find solutions that are of comparable size to those by MH and SIHC, but are better in scalability.

\begin{figure}[htbp]
\centering
\subfloat[ ]{\includegraphics[width=0.8\columnwidth]{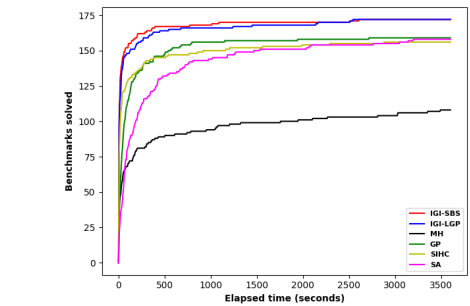}%
\label{fig-sliatime}}
\hspace{20pt}
\subfloat[ ]{\includegraphics[width=0.8\columnwidth]{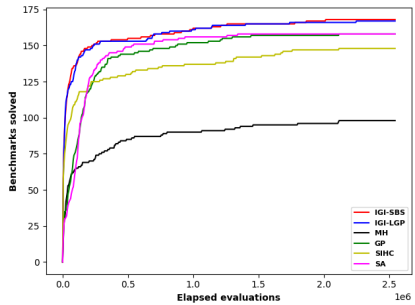}%
\label{fig-sliaeval}}
\caption{(a) Number of benchmarks solved versus the computation time; (b) Number of benchmarks solved versus the number of evaluations in the string transformation domain. }
\label{fig-sliate}
\end{figure}

Figure \ref{fig-sliate} plots the number of benchmarks solved with the increasing of 
computational time (Figure \ref{fig-sliate}\subref{fig-sliatime}) and evaluations (Figure \ref{fig-sliate}\subref{fig-sliaeval}). 
Unlike in the list manipulation domain, the trajectories of IGI-SBS and IGI-LGP start steep and almost flatten very quickly,
indicating that most of benchmarks in this domain do not pose great challenge to IGI.

\begin{table}[htbp]
\centering
%\begin{threeparttable}
\caption{Results for 10 hard benchmarks (in the string transformation domain) that are randomly chosen, 
where $n$ shows the number of examples for each benchmark. The data indicates the number of successful runs out of 50. 
Best performance is shown in bold. }
\label{tab-sliarepeat}  \scriptsize %\tabcolsep =10pt
\begin{ruledtabular}
\begin{tabular}{lccccccc}
%\toprule

Benchmarks\footnote{\scriptsize The full name of some benchmarks are given below.}  & $n$ &  IGI-SBS & IGI-LGP & MH & GP & SIHC & SA \\
 \colrule

31753108  &  3    &  \textbf{50} & \textbf{50} & 0 & 28 & 1 & 22 \\
44789427  &  4    &  \textbf{50} & \textbf{50} & 0 & 35 & 12 & 42 \\
exceljet2  &    3    &  \textbf{50} & \textbf{50} & 0 & 39 & 9 & 18 \\
enwfts\footnote{\scriptsize extract-nth-word-from-text-string}     &     4   &  \textbf{44} & 36 & 0 & 14 & 0 & 23 \\
ewtbwsc\footnote{\scriptsize extract-word-that-begins-with-specific-character}  &    3    &  \textbf{50} & 49 & 0 & 47 & 6 & 10 \\
gmnffn\footnote{\scriptsize get-middle-name-from-full-name}   &    4       & \textbf{50} & 49 & 0 & 22 & 0 & 18 \\
p10lr\footnote{\scriptsize phone-10-long-repeat}       &     400    & \textbf{46} & 42 & 9 & 26 & 18 & 34 \\
stsasc\footnote{\scriptsize split-text-string-at-specific-character}    &  4       & \textbf{50} & \textbf{50} & 0 & 43 & 1 & 31 \\
sncfc\footnote{\scriptsize strip-numeric-characters-from-cell}     &     3    &46 & 45 & 0 & \textbf{48} & 10 & 41 \\
univ\_4  & 8       & \textbf{49} & 42 & 0 & 32 & 1 & 27 \\

%\bottomrule
\end{tabular}
\end{ruledtabular}
%\begin{tablenotes}
%\item The full name of some benchmarks are as follows. enwfts: extract-nth-word-from-text-string; 
%ewtbwsc: extract-word-that-begins-with-specific-character;
%gmnffn: get-middle-name-from-full-name; p10lr: phone-10-long-repeat;
%stsasc: split-text-string-at-specific-character; sncfc: strip-numeric-characters-from-cell.
%\end{tablenotes}
%\end{threeparttable}
\end{table}

To further examine the performance of IGI, we select 10 hard benchmarks in the same way as 
we do in the list manipulation domain. Table \ref{tab-sliarepeat} shows the success rates for each algorithm 
on these 10 benchmarks. From Table \ref{tab-sliarepeat}, IGI-SBS always succeeds in  
6 benchmarks, while IGI-LGP always succeeds in 4. Both IGI-SBS and IGI-LGP achieve higher success rates than all the baselines on 
all the benchmarks except on sncfc where GP is slightly better. 
MH performs very poorly, which always fails in 9 out of 10 benchmarks. 
It seems that although SIHC is very competitive to GP and SA in terms of 
the number of benchmarks solved, it generally performs worse than GP and SA 
on hard benchmarks. The similar phenomenon
can be observed in the list manipulation domain. 
This implies that it is more difficult for the simple search mechanism of SIHC to adapt to 
hard synthesis tasks that may have complex fitness landscapes. 

In summary, IGI algorithms perform better than
all the other algorithms in terms of scalability in the string transformation domain, without sacrificing the
solution quality. In addition, they spend less average time to find solution programs,  
and also have much stronger ability to address hard synthesis tasks.

\subsection{On the Issue of Generalization}
\label{sec-On the Issue of Generalization}

Although the issue of generalization is not the focus of this paper,  
it is interesting to see how well the solution programs found by 
IGI algorithms generalize to unseen input-output examples. 
We generated 100 additional input-output examples for each of the 
200 benchmarks in the list manipulation domain using the corresponding oracle programs. 
Note that this cannot be done in the string transformation domain since  oracle programs
are unknown for those benchmarks. 

\begin{table}[htbp]
\caption{Percentage of solution programs that can generalize to 100 hold-out input-output examples. 
}\label{tab-gen} \scriptsize %\tabcolsep = 5.0pt
\begin{center}
\begin{ruledtabular}
\begin{tabular}{lcccccc}
%\toprule

Statistics & IGI-SBS & IGI-LGP & MH & GP & SIHC & SA \\
   \midrule
   
Total solved 	    							& 141 & 136       & 24     & 65    & 80 & 62 \\
Generalizable                                                              & 133  &  122     &   23   & 57    & 68 &   51                       \\
%\cmidrule{1-7}
\colrule
Percentage (\%)  &     94.33   & 89.71 & 95.83 & 87.69  & 85  & 82.26\\
%\bottomrule
\end{tabular}
\end{ruledtabular}
\end{center}
\end{table}

Table \ref{tab-gen} reports the percentage of solution programs that can generalize to 100 hold-out input-output examples for each considered algorithm. 
It can be seen that the IGI algorithms have better generalization ability than GP, SIHC and SA. 
This also implies that the better scalability of IGI is not due to  overfitting on given input-output examples. 
MH achieves the highest generalization percentage, but from a small basis as it only solves 24 relatively easy tasks 
with overfitting less likely to occur. 
It is worth mentioning that in our experiments all algorithms considered stop searching once a solution is found.
We can further alleviate overfitting by letting the algorithm return all solution programs found within the 
time budget and then choose the smallest one among them.

\section{Related Work}
\label{sec-Related Work}

Program synthesis is an active research topic including a large and diverse body of work, such as 
enumerative program synthesis \cite{alur2017scaling,lee2018accelerating,huang2020reconciling,lee2021combining}, 
constraint-based program synthesis \cite{solar2008program,srivastava2010program, jha2010oracle} and neural program synthesis \cite{devlin2017robustfill, parisotto2017neuro, hong2021latent,chen2021evaluating}. 
In what follows, we review some prior work on stochastic program synthesis that is most closely related to our proposed IGI.

\textbf{Metropolis-Hastings algorithm.} 
This line of research started from STOKE \cite{schkufza2013stochastic}, which tackles superoptimization problems
using a stochastic search strategy known as Metropolis-Hastings (MH). Due to the surprisingly good 
performance of STOKE on superoptimization, it had stirred great interest with its publication and was 
regarded as a timely and significant contribution \cite{gulwani2016technical} to program synthesis at that time. 
Alur et al. \cite{alur2013syntax} adapted STOKE and introduced a MH approach over programs represented by trees. 
This MH approach participated in SyGuS competitions 2014--2016 \cite{sygus2014}, but it did not achieve 
very competitive performance compared to other categories of techniques. Since then, the enthusiasm of researchers 
for stochastic synthesis seemed to drop and there was no stochastic solver participating in SyGuS competitions 
in the following years. Recently, several studies have been conducted to enhance the performance of STOKE. 
For example, Bunel et al. \cite{bunel2017learning} proposed to learn the proposal distribution in STOKE parameterized by a neural network, which is expected to better exploit the power of MH; Koenig et al. 
\cite{koenig2021adaptive} proposed an effective adaptive restart algorithm for addressing a limitation in STOKE where
the search often progresses via a series of plateaus. 
However, these enhancements have mostly targeted superoptimization, and it is still unclear how to make them amenable 
to  stochastic search over trees.

\textbf{Genetic programming.}
Genetic programming (GP) \cite{koza1992genetic,banzhaf1998genetic,poli2008field} has a much longer history than Metropolis-Hastings algorithm
for the purpose of program synthesis. Indeed, achieving automatic programming is arguably the most aspirational goal in the field of
GP \cite{o2019automatic}. In canonical GP, the variation operations are implemented via crossover and mutation. 
Borrowing ideas from estimation of distribution algorithms (EDAs) \cite{larranaga2001estimation}, 
an interesting alternative is to replace such variation operations with the process of sampling from a probability distribution \cite{salustowicz1997probabilistic,sastry2003probabilistic,yanai2003estimation}. 
One typical example in this family of GP is probabilistic incremental program evolution (PIPE) \cite{salustowicz1997probabilistic}, where
the population is replaced by a hierarchy of probability tables with the tree structure. 
Although EDA-based GP approaches are appealing, they usually fail to offer significant performance gains over
standard tree-based GP \cite{poli2008field}, which remains to be investigated. 
It is known that the most widespread type of GP expresses programs as syntax trees, but there are other GP based
program synthesizers which use different program representations. 
PushGP \cite{spector2002genetic} and grammar-guided GP (GGGP) \cite{mckay2010grammar} are among the most representative ones. 
PushGP evolves programs expressed in the Push programming language, which supports different data types by 
providing a stack for each data type as well as for the code that is executed. 
Recent studies around PushGP focus on designing new 
mutation operators \cite{helmuth2018program} and parent selection methods \cite{helmuth2016lexicase, helmuth2020benchmarking} in order to further improve its performance.    
GGGP can either use the derivation tree \cite{whigham1995grammatically} or a linear genome \cite{o2001grammatical} as its program representation, 
which can be mapped to a resulting program via the context-free grammar. 
Recently, some work on GGGP has paid attention to 
grammar design \cite{forstenlechner2017grammar}, generalizability \cite{sobania2021generalizability} and the quality of generated code \cite{sobania2019teaching, hemberg2019domain}.

\textbf{Stochastic local search.} Although stochastic local search (SLS) \cite{hoos2004stochastic}
has been used extensively for combinatorial problems, there is surprisingly few research on SLS for 
program synthesis. The most notable work in this direction is O'Reilly's PhD thesis \cite{o1995analysis}, where 
stochastic iterated hill climbing (SIHC) and simulated annealing (SA) were investigated for solving 
program discovery problems. Her results indicate that SIHC and SA are generally comparable 
to GP and even sometimes outperform GP. In addition to this work, there are some other research efforts on SLS that 
target a specific synthesis problem. 
For example, Nguyen et al. \cite{nguyen2014automatic} proposed to evolve dispatching rules
in job-sop scheduling via iterated local search; Kantor et al.
\cite{kantor2021simulated} presented simulated annealing using the interaction-transformation representation
for symbolic regression.

But all these methods have not (yet) succeeded in bringing into being a comprehensive methodology for program synthesis. So it remains to be seen which technique has the most power and will be adopted by the wider community.

\section{Conclusion}
\label{sec-Conclusion}

In this paper, we have proposed IGI, a new framework for 
stochastic program synthesis.
Inspired by the process of human programming, IGI considers a sequence of
program improvements by iteratively searching for modifications to the current reference program.
In terms of solution representation, IGI uses a differential code representation that will undergo epochs of evolution under
the control of input-output examples. IGI can therefore also be seen as a kind of generative or developmental
genetic programming \cite{koza2010human,eiben2015evolutionary}, which allows the reuse of code and helps to scale up the complexity of 
evolved programs.  
Experimental results on two different application domains have demonstrated 
the clear advantage of IGI over several representative SPS techniques in terms of scalability and solution quality. 

In the future, we will investigate how to extend IGI to popular general-purpose
programming languages \cite{chen2021evaluating}.
It would also be interesting to use IGI as the symbolic search engine in neurosymbolic programming \cite{chaudhuri2021neurosymbolic}.

% Our IGI approach can 
% address the scalability issue of SPS techniques in an elegant way

\vspace{3mm}
\begin{acknowledgements}
Authors gratefully acknowledge support from the Koza Endowment fund to Michigan State University. Runs were done on MSU's iCER HPCC system. 
\end{acknowledgements}

\bibliographystyle{spmpsci-modi}
\bibliography{reference}

%% The Appendices part is started with the command \appendix;
%% appendix sections are then done as normal sections
\clearpage
\onecolumngrid

\appendix

\section{DSL for List Manipulation (DSL-LM)}
\label{sec-DSL for List Manipulation}

This DSL is introduced in the DeepCoder paper \cite{balog2017deepcoder}, which is
inspired by query languages such as SQL.
It contains a number of high-level functions that can be
used to manipulate integer arrays or integers. In Tables \ref{tab-dsllm} and \ref{tab-dsllm2} 
we list all of the functions of DSL-LM and give their symbols, 
argument types, return types and detailed descriptions.  For each benchmark task in this domain,
all these functions are available. 
Note that there are no constants in DSL-LM, and terminals are all from the 
program's external inputs. 

\begin{table}[htbp]
\caption{Description of functions in the DSL-LM.
}\label{tab-dsllm} \scriptsize %\tabcolsep = 8.0pt
\begin{center}
\begin{ruledtabular}
\begin{tabular}{clll}
%\toprule
    Symbol &   Arguments  & Return Type & Description \\
\colrule

\texttt{HEAD}  &  $x$: Integer array  &   Integer  &  Return the first element of a given array $x$ (or \\
			&&&  NULL if $x$ is empty). \\
\cmidrule{1-4}

\texttt{LAST}  &  $x$: Integer array  &   Integer  &  Return the last element of a given array $x$ (or   \\
			&&& NULL if $x$ is empty). \\
\cmidrule{1-4}
\texttt{TAKE}  & $n$: Integer;    & Integer array & Given an integer $n$ and an array $x$, return the   \\
		    & $x$: Integer array & & array truncated after the $n$-th element. (If the  \\
		    & & & length of $x$ is not larger than $n$, return $x$ without \\
		    &&&  modification.) \\
 
 \cmidrule{1-4}

\texttt{DROP}  & $n$: Integer;    & Integer array & Given an integer $n$ and an array $x$, return the \\
		    & $x$: Integer array &&  array with the first $n$ elements dropped. (If the  \\
		    	&&&   length of x is not larger than $n$, return an empty\\
 			&&&		 array.) \\
\cmidrule{1-4}

\texttt{ACCESS}  & $n$: Integer;    & Integer& Given an integer $n$ and an array $x$, return the \\
		    & $x$: Integer array  &&  ($n+1$)-th element of $x$. (If the length of $x$ is not\\
		                                    &&&   larger than $n$, return NULL.) \\
\cmidrule{1-4} 

\texttt{MINIMUM}  &  $x$: Integer array & Integer & Return the minimum of a given array (or NULL \\
							  &&&       if $x$ is empty).   \\
\cmidrule{1-4}

\texttt{MAXIMUM}  &  $x$: Integer array & Integer &  Return the maximum of a given array (or NULL \\
							 &&&        if $x$ is empty).   \\

\cmidrule{1-4}

\texttt{REVERSE}  &  $x$: Integer array & Integer array & Return the elements of a given array $x$ in reversed \\
											&&& order.  \\
\cmidrule{1-4}

\texttt{SORT}  &  $x$: Integer array & Integer array   &  Return the elements of a given array $x$ in  \\
						&&&  non-decreasing order.\\
\cmidrule{1-4}

\texttt{SUM}  &  $x$: Integer array & Integer &  Return the sum of the elements in a given array $x$.\\
\cmidrule{1-4}

\texttt{MAPA1}  &  $x$: Integer array & Integer array  &   Each element in a given array $x$ plus 1, and the\\
										&&& modified array is returned.     \\
\cmidrule{1-4}

\texttt{MAPM1}  &  $x$: Integer array & Integer array  &   Each element in a given array $x$ minus 1, and the\\
										&&& modified array is returned.     \\
\cmidrule{1-4}

\texttt{MAPT2}  &  $x$: Integer array & Integer array  &   Each element in a given array $x$ is multiplied by 2, \\
										&&& and the modified array is returned.     \\
\cmidrule{1-4}

\texttt{MAPT3}  &  $x$: Integer array & Integer array  &   Each element in a given array $x$ is multiplied by 3, \\
										&&& and the modified array is returned.     \\
\cmidrule{1-4}

\texttt{MAPT4}  &  $x$: Integer array & Integer array  &   Each element in a given array $x$ is multiplied by 4, \\
										&&& and the modified array is returned.     \\
\cmidrule{1-4}

\texttt{MAPD2}  &  $x$: Integer array & Integer array &  Each element in a given array $x$ divided by 2 (fractions  \\
										&&&  are rounded down), and the modified array is returned.     \\
\cmidrule{1-4}

\texttt{MAPD3}  &  $x$: Integer array & Integer array &  Each element in a given array $x$ divided by 3 (fractions  \\
										&&&are rounded down), and the modified array is returned.     \\
\cmidrule{1-4}

\texttt{MAPD4}  &  $x$: Integer array & Integer array &  Each element in a given array $x$ divided by 4 (fractions  \\
										&&& are rounded down), and the modified array is returned.     \\
\cmidrule{1-4}

\texttt{MAPV1}  &  $x$: Integer array & Integer array  &   Each element in a given array $x$ is multiplied by -1, \\
										&&& and the modified array is returned.     \\
\cmidrule{1-4}

\texttt{MAPP2}  &  $x$: Integer array & Integer array  &   Each element in a given array $x$ is multiplied by \\
										&&& itself, and the modified array is returned.     \\
\cmidrule{1-4}

\texttt{FILG0}  &  $x$: Integer array & Integer array   & Return the elements of a given array $x$ that are larger \\
&&& than 0 in their original order.  \\
\cmidrule{1-4}

\texttt{FILL0}  &  $x$: Integer array & Integer array   & Return the elements of a given array $x$ that are less \\
&&&  than 0 in their original order.  \\
\cmidrule{1-4}

\texttt{FILEV}  &  $x$: Integer array & Integer array   & Return the elements of a given array $x$ that are even \\
&&&  in their original order.  \\
\cmidrule{1-4}

\texttt{FILOD}  &  $x$: Integer array & Integer array   & Return the elements of a given array $x$ that are odd \\
&&&  in their original order.  \\
\cmidrule{1-4}

\texttt{COUG0}  &  $x$: Integer array & Integer  & Return the number of elements in a given array $x$ \\
									&&& that is larger than 0. \\
\cmidrule{1-4}

\texttt{COUL0}  &  $x$: Integer array & Integer  & Return the number of elements in a given array $x$  \\
									&&& that is less than 0. \\
\cmidrule{1-4}

\texttt{COUEV}  &  $x$: Integer array & Integer  & Return the number of elements in a given array $x$ \\
                              &&&  that is even.  \\
\cmidrule{1-4}

\texttt{COUOD}  &  $x$: Integer array & Integer  & Return the number of elements in a given array $x$  \\
			&&& that is odd.\\
\cmidrule{1-4}

%\bottomrule
\end{tabular}
\end{ruledtabular}
\end{center}
\end{table}

\begin{table}[htbp]
\caption{Description of functions in the DSL-LM (continued).
}\label{tab-dsllm2} \scriptsize %\tabcolsep = 8.0pt
\begin{center}
\begin{ruledtabular}
\begin{tabular}{clll}
\toprule
    Symbol &   Arguments  & Return Type & Description \\
\colrule

\texttt{ZIPSUM}  &  $x$: Integer array & Integer array  & Return an array $z$ with length $n$ where \\
			&  $y$: Integer array  &&   $z[i]=x[i]+y[i]$, $i=0,1,\ldots n-1$ and $n$ is \\
			&&&  the minimum of the lengths of $x$ and $y$. \\ 
\cmidrule{1-4}

\texttt{ZIPDIF}  &  $x$: Integer array & Integer array  & Return an array $z$ with length $n$ where \\
			&  $y$: Integer array  &&   $z[i]=x[i]-y[i]$, $i=0,1,\ldots n-1$ and $n$ is \\
			&&&  the minimum of the lengths of $x$ and $y$. \\ 
\cmidrule{1-4}

\texttt{ZIPMUL}  &  $x$: Integer array & Integer array  & Return an array $z$ with length $n$ where \\
			&  $y$: Integer array  &&   $z[i]=x[i] * y[i]$, $i=0,1,\ldots n-1$ and $n$ is \\
			&&&  the minimum of the lengths of $x$ and $y$. \\ 
\cmidrule{1-4}

\texttt{ZIPMAX}  &  $x$: Integer array & Integer array  & Return an array $z$ with length $n$ where \\
			&  $y$: Integer array  &&   $z[i]=\max\{x[i], y[i]\}$, $i=0,1,\ldots n-1$ \\
			&&&  and $n$ is the minimum of the lengths of $x$ and $y$. \\ 
\cmidrule{1-4}

\texttt{ZIPMIN}  &  $x$: Integer array & Integer array  & Return an array $z$ with length $n$ where \\
			&  $y$: Integer array  &&   $z[i]=\min\{x[i], y[i]\}$, $i=0,1,\ldots n-1$ \\
			&&&  and $n$ is the minimum of the lengths of $x$ and $y$. \\ 

\cmidrule{1-4}
	
\texttt{SCANSUM}  &  $x$: Integer array & Integer array  &  Returns an array $y$ of the same length as $x$ and \\
&&&    with its content defined by the recurrence:  \\
&&& $y[0]=x[0]$, $y[i]=y[i-1]+x[i]$, for $i \geq 1$.\\
\cmidrule{1-4}

\texttt{SCANDIF}  &  $x$: Integer array & Integer array  &  Returns an array $y$ of the same length as $x$ and \\
&&&    with its content defined by the recurrence:  \\
&&& $y[0]=x[0]$, $y[i]=y[i-1]-x[i]$, for $i \geq 1$.\\
\cmidrule{1-4}

\texttt{SCANMUL}  &  $x$: Integer array & Integer array  &  Returns an array $y$ of the same length as $x$ and \\
&&&    with its content defined by the recurrence:  \\
&&& $y[0]=x[0]$, $y[i]=y[i-1]*x[i]$, for $i \geq 1$.\\
\cmidrule{1-4}

\texttt{SCANMAX}  &  $x$: Integer array & Integer array  &  Returns an array $y$ of the same length as $x$ and \\
&&&    with its content defined by the recurrence:  \\
&&& $y[0]=x[0]$, $y[i]=\max\{y[i-1], x[i]\}$, for $i \geq 1$.\\
\cmidrule{1-4}

\texttt{SCANMIN}  &  $x$: Integer array & Integer array  &  Returns an array $y$ of the same length as $x$ and \\
&&&    with its content defined by the recurrence:  \\
&&& $y[0]=x[0]$, $y[i]=\min\{y[i-1], x[i]\}$, for $i \geq 1$.\\
			
\bottomrule
\end{tabular}
\end{ruledtabular}
\end{center}
\end{table}

\clearpage

\section{DSL for String Transformation (DSL-ST)}
\label{sec-DSL for String Transformation}

This DSL is designed for the PBE-Strings track in the SyGuS competition \cite{sygus2014}. 
Table \ref{tab-dslst} describes all of the functions of DSL-ST in detail. 
Terminals include constants and the program's external inputs.
For each task of the SyGuS benchmarks 
the definition file specifies -- besides the input-output examples -- what functions in Table \ref{tab-dslst} are used 
and provides some string, integer and Boolean constants.

\begin{table}[htbp]
\caption{Description of functions in the DSL-ST.
}\label{tab-dslst} \scriptsize %\tabcolsep = 8.0pt
\begin{center}
\begin{ruledtabular}
\begin{tabular}{clcl}
\toprule
    Symbol &   Arguments  & Return Type & Description \\
\colrule

\texttt{CAT}  &  $s$: String & String  &  Concatenate the strings $s$ and $t$, and return \\
		    & $t$: String &              &  the combined string.     \\

\cmidrule{1-4}

\texttt{REP}  &  $s$: String    & String  &  Return a copy of the string $s$ where the first \\
		    &  $t$: String    &            &   occurrence of a substring $t$ is replaced with       \\
		    &  $r$: String    &            &     another substring $r$.   \\
\cmidrule{1-4}
	    
\texttt{AT}  &  $s$: String    & String  &  Return the string containing a single character   \\
		    &  $i$: Integer    &            &  at index $i$ in $s$, i.e., $s[i]$. (If $i<0$ or $i$ is no smaller     \\
		    					&&& than the length of $s$, return an empty string.)  \\

\cmidrule{1-4}

\texttt{ITS}  &  $x$: Integer   & String  & If integer $x$ is not smaller than 0, convert $x$ into   \\
                                                         &&&  the string and return the string. Otherwise, \\
                                                         &&&  return an empty string. \\
\cmidrule{1-4}

\texttt{SITE}  &  $b$: Boolean   & String  &   If $b$ is true, return $s$, otherwise return $t$. \\
		   &    $s$: String      &   \\
		   &  $t$: String      &     \\
\cmidrule{1-4}

\texttt{SUBSTR}  &  $s$: String   & String  &   Return the substring of a given string $s$ with the \\
		   &    $i$: Integer      && start index $i$ and the end index $\min\{n,i + j \}-1$,   \\
		   &    $j$: Integer      &&  where $n$ is the length of $s$. (If $i<0$ or $j<0$ or    \\
		   				&&& $i\geq n$, return an empty string.)    \\
\cmidrule{1-4}

\texttt{ADD}  &  $x$: Integer    & Integer &  Return the sum of integers $x$ and $y$. \\
		    &  $y$: Integer    &            &         \\
\cmidrule{1-4}

\texttt{SUB}  &  $x$: Integer    & Integer &  Return the difference between integers $x$ and $y$. \\
		    &  $y$: Integer    &            &         \\

\cmidrule{1-4}

\texttt{LEN}  &  $s$: String   & Integer &  Return the length of a given string $s$.\\
\cmidrule{1-4}

\texttt{STI}  &  $s$: String   & Integer & If all characters in the string $s$ are digits, convert \\
						&&& $s$ into the integer and return the integer.  \\
						&&&  Otherwise return $-1$.  \\
\cmidrule{1-4}

\texttt{IITE}  &  $b$: Boolean   & Integer &   If $b$ is true, return $x$, otherwise return $y$. \\
		   &    $x$: Integer      &   \\
		   &  $y$: Integer      &     \\
\cmidrule{1-4}

\texttt{IND}  &  $s$: String   & Integer &  Search the string $t$ in the substring of $s$ that starts at \\
		   &  $t$: String      &&  index $i$ and ends at index $n-1$, where $n$ is the length   \\
		   &  $i$: Integer      &&   of $s$, and return the lowest index in $s$ where $t$ is found.   \\
		                    &&&  (If $i<0$ or $i \geq n$ or $t$ is not found in $s$, return $-1$.)  \\
\cmidrule{1-4}

\texttt{EQ}  &  $x$: Integer   & Boolean &   If $x$ equals to $y$, return true, otherwise return\\
		   & $y$: Integer      &&   false. \\
\cmidrule{1-4}

\texttt{PRF}  & $s$: String  & Boolean & If $s$ is the prefix of $t$, return true, otherwise return   \\
		   &  $t$: String     && 						false.  \\
\cmidrule{1-4}

\texttt{SUF}  & $s$: String  & Boolean &  If $s$ is the suffix of $t$, return true, otherwise return  \\
		   &  $t$: String     &&  false. \\
\cmidrule{1-4}

\texttt{CONT}  & $s$: String  & Boolean &   If $t$ is found in $s$, return true, otherwise return\\
		   &  $t$: String     &&  false. \\
\bottomrule
\end{tabular}
\end{ruledtabular}
\end{center}
\end{table}

\clearpage

\twocolumngrid
\section{Parameter Selection}
\label{sec-Parameter Selection}

Key parameters of each algorithm were tuned by performing grid search on two hard synthesis tasks
from the list and string transformation domains. 
For each parameter combination of an algorithm, we performed 10 independent runs 
using a timeout criterion of 20 minutes on the two tasks. 
%Then for each of the two tasks, 
We ranked all parameter combinations of an algorithm 
according to the average maximum fitness achieved over 10 runs and
selected the parameter combination with the lowest average rank over the two tasks. 
For IGI-SBS, the space of parameters considered is beam width $\in \{ 25, 50, 100\}$,  
number of successors for each patch is $\in \{5, 10, 20\}$, the maximum length of patches considered is $\in \{2, 3, 4 \}$,
and tournament size is $\in \{ 2, 5, 8 \}$, for a total of 81 combinations. 
For IGI-LGP, the space of parameters considered is population size $\in \{50 * i | i =1, \ldots, 8 \}$, 
maximum number of generations in each epoch is
$\in \{5, 10, 20 \}$
and tournament size is $\in \{ 2, 5, 8 \}$ for a total of 72 combinations.
For MH, the space of parameters considered is switch probability $\in \{0.002 * i  | i=1, 2, \ldots, 10\}$
and smoothing constant $\in \{0.1, 0.3, 0.5, 0.7, 0.9\}$ for a total of 50 combinations. 
For GP, the space of parameters considered is population size $\in \{5000, 10000, 20000\}$, 
crossover probability $\in \{0.8, 0.9, 1.0\}$,  
mutation probability (per node) $\in \{0.05, 0.08, 0.1\}$ and
tournament size $\in \{ 2, 5, 8 \}$ for a total of 81 combinations. 
For SIHC, the space of parameters considered is the maximum number of mutations $\in\{500 * i | i =1, 2, \ldots, 40\}$ for a total of 40 combinations. 
For SA, the space of parameters considered is final temperature $\in \{0.0001, 0.0005, 0.001, 0.005, 0.01\}$ and stepsize  
$\in \{500*i |i=1,2,\ldots10\}$ for a total of 50 combinations.

The final tuned parameters for all algorithms considered are shown in Tables \ref{tab-paramigisbs} to
%\ref{tab-paramigilgp}, \ref{tab-parammh}, \ref{tab-paramgp}, \ref{tab-paramihc} and 
\ref{tab-paramsa}. 
These parameters are used throughout our experiments.

\vspace{2cm}

\begin{table}[htbp]
\caption{Parameter setting of IGI-SBS. 
}\label{tab-paramigisbs} \scriptsize %\tabcolsep = 10.0pt
\begin{center}
\begin{ruledtabular}
\begin{tabular}{lc}
\toprule
Parameter &  Value \\
\colrule
Beam width ($B$)    &  50 \\ 
Number of successors of each patch ($C$)                  	& 5   \\
Maximum length of patches considered ($L_{\max}$)   	& 3  \\
Tournament size & 2  \\
Number of initial programs ($K$)   				& $B \times C \times L_{\max}$    \\
Number of perturbations ($M$)                     		& 200                 \\
Minimum perturbation strength ($S_{\min}$) 			& 4                          \\
Minimum depth of initial programs ($d_{\min}$) 			& 2 				\\
Maximum depth of initial programs ($d_{\max}$) 		& 4 				\\

\bottomrule
\end{tabular}
\end{ruledtabular}
\end{center}
\end{table}

\begin{table}[htbp]
\renewcommand{\arraystretch}{1.0}
\caption{Parameter setting of IGI-LGP. 
}\label{tab-paramigilgp} \scriptsize \tabcolsep = 12.0pt
\begin{center}
\begin{ruledtabular}
\begin{tabular}{lc}
\toprule
Parameter &  Value \\
\colrule
Population size ($N$) & 100 \\
Maximum generations ($G$)  & 5  \\
Tournament size & 2  \\
Crossover probability  & 1.0 \\
Mutation probability & 1.0 \\
Number of initial programs ($K$)   				& $N \times G$    \\
Number of perturbations ($M$)                     		& 200                 \\
Minimum perturbation strength ($S_{\min}$) 		& 4                          \\
Minimum depth of initial programs ($d_{\min}$) 		& 2 				\\
Maximum depth of initial programs ($d_{\max}$) 	& 4 				\\
\bottomrule
\end{tabular}
\end{ruledtabular}
\end{center}
\end{table}

\begin{table}[htbp]
\renewcommand{\arraystretch}{1.0}
\caption{Parameter setting of MH. 
}\label{tab-parammh} \scriptsize \tabcolsep = 12.0pt
\begin{center}
\begin{ruledtabular}
\begin{tabular}{lc}
\toprule
Parameter &  Value \\
\colrule
Switch probability ($p_{m}$)  & 0.006 \\
Smoothing constant ($\beta$) & 0.7 \\
Maximum allowed depth of programs & 30 \\
\bottomrule
\end{tabular}
\end{ruledtabular}
\end{center}
\end{table}

\begin{table}[htbp]
\renewcommand{\arraystretch}{1.0}
\caption{Parameter setting of GP. 
}\label{tab-paramgp} \scriptsize \tabcolsep = 12.0pt
\begin{center}
\begin{ruledtabular}
\begin{tabular}{lc}
\toprule
Parameter &  Value \\
\colrule
Population size  & 20000 \\
Crossover probability & 0.9 \\
Mutation probability (per node)  &  0.1 \\
Tournament size & 2  \\
Minimum depth of initial programs ($d_{\min}$) & 2 \\
Maximum depth of initial programs ($d_{\max}$)   & 4 \\
Maximum allowed depth of programs & 30 \\
\bottomrule
\end{tabular}
\end{ruledtabular}
\end{center}
\end{table}

\begin{table}[htbp]
\renewcommand{\arraystretch}{1.0}
\caption{Parameter setting of SIHC. 
}\label{tab-paramihc} \scriptsize \tabcolsep = 12.0pt
\begin{center}
\begin{ruledtabular}
\begin{tabular}{lc}
\toprule
Parameter &  Value \\
\colrule
Maximum number of mutations ($T_{\max}$)  &  500 \\
Minimum depth of initial programs ($d_{\min}$)  & 2 \\
Maximum depth of initial programs ($d_{\max}$)   & 4 \\
Maximum allowed depth of programs & 30 \\
\bottomrule
\end{tabular}
\end{ruledtabular}
\end{center}
\end{table}

\begin{table}[htbp]
\renewcommand{\arraystretch}{1.0}
\caption{Parameter setting of SA. 
}\label{tab-paramsa} \scriptsize \tabcolsep = 10.0pt
\begin{center}
\begin{ruledtabular}
\begin{tabular}{lc}
\toprule
Parameter &  Value \\
\colrule
Starting temperature   & 1.5  \\
Final temperature &     0.001 \\
Stepsize   &  500 \\
Minimum depth of initial programs ($d_{\min}$) & 2 \\
Maximum depth of initial programs ($d_{\max}$)  & 4 \\
Maximum allowed depth of programs & 30 \\

\bottomrule
\end{tabular}
\end{ruledtabular}
\end{center}
\end{table}

\clearpage

\end{document}